\documentclass[conference]{IEEEtran}
\IEEEoverridecommandlockouts
\usepackage{cite}
\usepackage{amsmath,amssymb,amsfonts}
\usepackage{algorithmic}
\usepackage{graphicx}
\usepackage{textcomp}
\usepackage{xcolor}
\usepackage{url}
\usepackage{multirow} 
\usepackage{tabularx}
\usepackage{booktabs}      
\def\BibTeX{{\rm B\kern-.05em{\sc i\kern-.025em b}\kern-.08em
    T\kern-.1667em\lower.7ex\hbox{E}\kern-.125emX}}
\begin{document}

\title{SEED: A Structural Encoder for Embedding-Driven Decoding in Time Series Prediction with LLMs}

\author{
\IEEEauthorblockN{Fengze Li}
\IEEEauthorblockA{
\textit{Xi'an Jiaotong-Liverpool University} \\
\textit{University of Liverpool} \\
Suzhou, China\\
FengzeLi@liverpool.ac.uk}
\and
\IEEEauthorblockN{Yue Wang}
\IEEEauthorblockA{
\textit{Xi'an Jiaotong-Liverpool University} \\
Suzhou, China\\
Yue.Wang2104@student.xjtlu.edu.cn}
\and
\IEEEauthorblockN{Yangle Liu}
\IEEEauthorblockA{
\textit{University of Liverpool} \\
\textit{Xi'an Jiaotong-Liverpool University} \\
Liverpool, UK\\
sgyli136@liverpool.ac.uk}
\and
\IEEEauthorblockN{Ming Huang}
\IEEEauthorblockA{
\textit{Xi'an Jiaotong-Liverpool University} \\
\textit{University of Liverpool} \\
Suzhou, China\\
Ming.Huang2202@student.xjtlu.edu.cn}
\and
\IEEEauthorblockN{Dou Hong}
\IEEEauthorblockA{
\textit{Xi'an Jiaotong-Liverpool University} \\
\textit{University of Liverpool} \\
Suzhou, China\\
Dou.Hong16@student.xjtlu.edu.cn}
\and
\IEEEauthorblockN{Jieming Ma\textsuperscript{*}}
\IEEEauthorblockA{
\textit{Xi'an Jiaotong-Liverpool University} \\
Suzhou, China \\
Jieming.Ma@xjtlu.edu.cn\\
\textsuperscript{*}Corresponding Author}
}

\maketitle

\begin{abstract}
Multivariate time series forecasting requires models to simultaneously capture variable-wise structural dependencies and generalize across diverse tasks. While structural encoders are effective in modeling feature interactions, they lack the capacity to support semantic-level reasoning or task adaptation. Conversely, large language models (LLMs) possess strong generalization capabilities but remain incompatible with raw time series inputs. This gap limits the development of unified, transferable prediction systems. Therefore, we introduce SEED, a structural encoder for embedding-driven decoding, which integrates four stages: a token-aware encoder for patch extraction, a projection module that aligns patches with language model embeddings, a semantic reprogramming mechanism that maps patches to task-aware prototypes, and a frozen language model for prediction. This modular architecture decouples representation learning from inference, enabling efficient alignment between numerical patterns and semantic reasoning. Empirical results demonstrate that the proposed method achieves consistent improvements over strong baselines, and comparative studies on various datasets confirm SEED's role in addressing the structural-semantic modeling gap.
\end{abstract}

\vspace{1em}


\section{Introduction}
\label{sec:intro}
Multivariate time series prediction is a fundamental task in temporal modeling, with broad applications in energy demand estimation, traffic management, industrial control, and financial prediction~\cite{mendis2024multivariate}. This task requires learning to predict future values from historical sequences involving multiple interdependent variables, often under shifting dynamics, missing values, and domain-specific noise~\cite{cai2024msgnet}. In real-world settings, forecasting systems are expected to support a variety of downstream tasks such as long-horizon prediction, imputation, and anomaly detection, all while being robust across data domains and temporal patterns~\cite{shao2024exploring}.

Traditional time series prediction models such as autoregressive–integrated moving average (ARIMA) and other statistics variants have long been applied for univariate and seasonal forecasting due to their statistical interpretability and low training cost~\cite{ray2023arima, chia2022long, kumar2022multi}. However, they struggle with nonlinear dynamics, high-dimensional multivariate settings, and long-range temporal dependencies. To overcome these limitations, deep learning models such as long short-term memory (LSTM)~\cite{lstm} and temporal convolution networks have been widely adopted. Although LSTM captures sequential dependencies through gated mechanisms, its reliance on recursive updates limits scalability and makes parallelization difficult. Temporal convolution networks offer better computational efficiency but are constrained by fixed receptive fields and lack global context awareness. 

Recent years have witnessed substantial progress with transformer-based architectures~\cite{vaswani2017attention, kitaev2020reformer}, which leverage attention mechanisms to model global dependencies in parallel. Beyond deterministic modeling, latent diffusion transformers~\cite{feng2024latent} introduce a generative approach for probabilistic forecasting by combining diffusion processes with attention-based temporal modeling, offering expressive uncertainty quantification. multi-resolution time-series transformers~\cite{zhang2024multi} propose a hierarchical encoder-decoder structure to capture long-term trends at different temporal resolutions. For detection tasks, inter-variable attention mechanisms~\cite{kang2024transformer} have proven effective in identifying anomalies that arise from complex correlations across variables. 

Furthermore, Informer~\cite{zhou2021informer} and Autoformer~\cite{wu2021autoformer} introduce sparse attention and series decomposition modules to improve efficiency and forecasting stability. Patch-based models like PatchTST~\cite{Yuqietal-2023-PatchTST} segment the input sequence into fixed-length patches and treat them as tokens, enabling scalable modeling of local temporal patterns. Also, iTransformer~\cite{liu2023itransformer} employs an inverted attention mechanism that shifts attention to the variable dimension, enhancing cross-variable structural encoding in multivariate settings. Despite their architectural diversity, these models rely on end-to-end training and are tailored to specific forecasting or detection tasks. They remain limited in semantic abstraction and require full model adaptation when deployed in new settings. This motivates frameworks that decouple structural encoding from downstream reasoning.

Recent advances leverage large language models (LLMs) for cross-domain generalization, and a new line of work draws from LLMs and prompt-based learning. PromptCast~\cite{xue2023promptcast} introduces prompt-conditioned forecasting with transformer decoders, allowing flexible task definition. It takes a different path by enabling task conditioning via natural language prompts, illustrating how frozen pretrained decoders can be directed to perform forecasting without retraining. Time-LLM~\cite{jin2023time} takes this further by reprogramming numeric time series patches into token embeddings interpretable by frozen LLMs, achieving zero-shot performance without fine-tuning. TimeGPT1~\cite{garza2023timegpt} presents a general foundation model pre-trained on massive time series corpora, showing strong performance in zero-shot and transfer forecasting. Nonetheless, these approaches often lack strong structural encoding on the input side and rely on complex, task-specific strategies to bridge numerical representations with semantic reasoning. This highlights the need for a unified framework that decouples structural encoding from semantic inference while supporting modular alignment with pretrained language models.

To address these limitations, we propose structural encoder for embedding-driven decoding (SEED), a modular framework designed to bridge the gap between structural representation and semantic reasoning by interfacing multivariate time series with pretrained LLMs. Therefore, the primary contributions of this work are as follows:
\begin{itemize}
\item A structural encoder for embedding-driven decoding (SEED) is introduced to address the structural-semantic gap in multivariate time series forecasting. The proposed framework decouples temporal representation learning from semantic inference, enabling modular integration with pretrained language models for prediction.

\item A token-aware structural encoder and a patch projection and alignment module are designed to capture variable-wise dependencies and transform numerical sequences into patch-level embeddings aligned with the input space of language models.

\item A semantic reprogramming module based on learnable prototypes and prompts, together with a frozen language model decoder, enables semantic token construction and downstream prediction without updating pretrained model parameters.

\item The effectiveness of each module is validated through experiments on forecasting tasks. Results from comparative studies on different datasets demonstrate the performance advantages of the proposed framework.
\end{itemize}

\begin{figure*}[t]
\centerline{\includegraphics[width=1\textwidth]{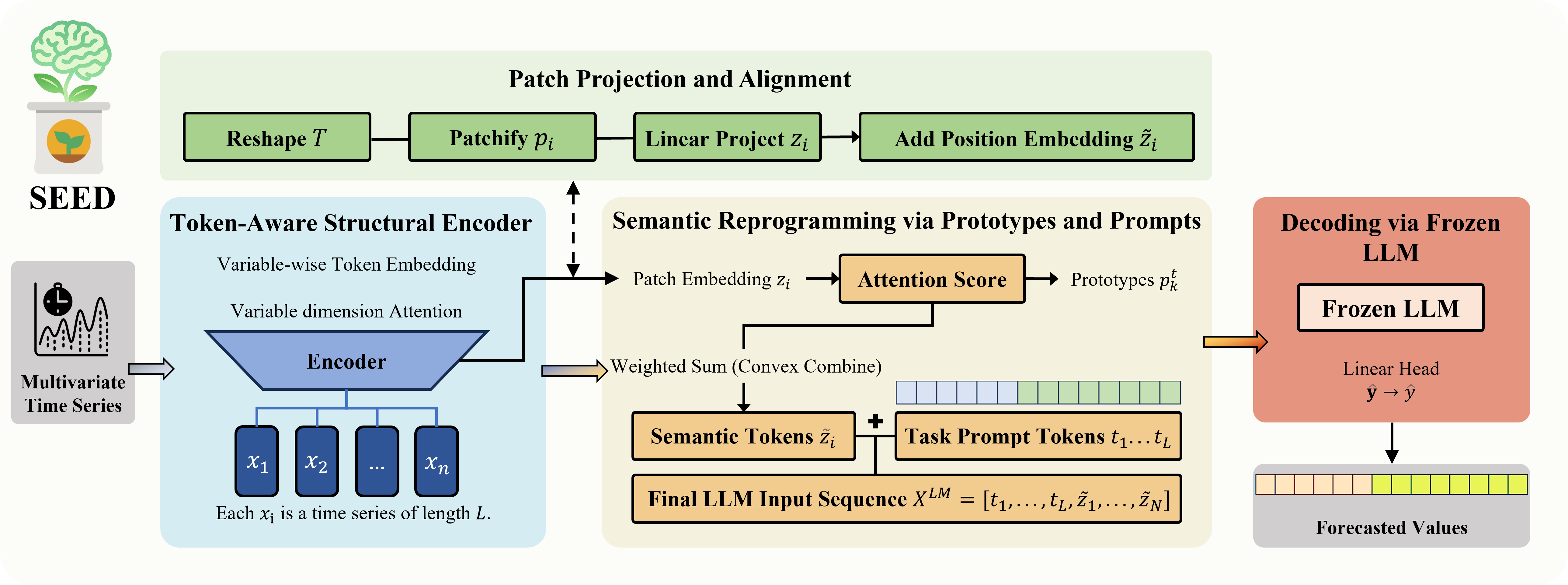}}
\caption{Overall architecture of SEED.}
\label{fig01}
\end{figure*}

\section{SEED}
The proposed structural encoder for embedding-driven decoding (SEED) is designed as a modular framework that decouples the structural encoding of multivariate time series from downstream semantic reasoning, shown as Fig. \ref{fig01}. The model is composed of four key components, each responsible for transforming raw time series data into interpretable, task-aligned representations suitable for LLM inference. The process begins with a token-aware structural encoder that captures inter-variable dependencies by inverting the conventional temporal tokenization scheme. 

First, a token-aware structural encoder models inter-variable dependencies by inverting the conventional temporal tokenization, treating each variable as a distinct token (Section~\ref{ss1}). The resulting representations are then segmented into fixed-length patches and projected into the embedding space expected by the target LLM (Section~\ref{ss2}). To bridge the gap between numerical features and language semantics, a semantic reprogramming module maps patch embeddings to interpretable token sequences using learnable prototypes and task-aware prompts (Section~\ref{ss3}). The resulting sequence is processed by a frozen autoregressive LLM for prediction (Section~\ref{ss4}).
\begin{figure}[thb]
\centerline{\includegraphics[width=1\columnwidth]{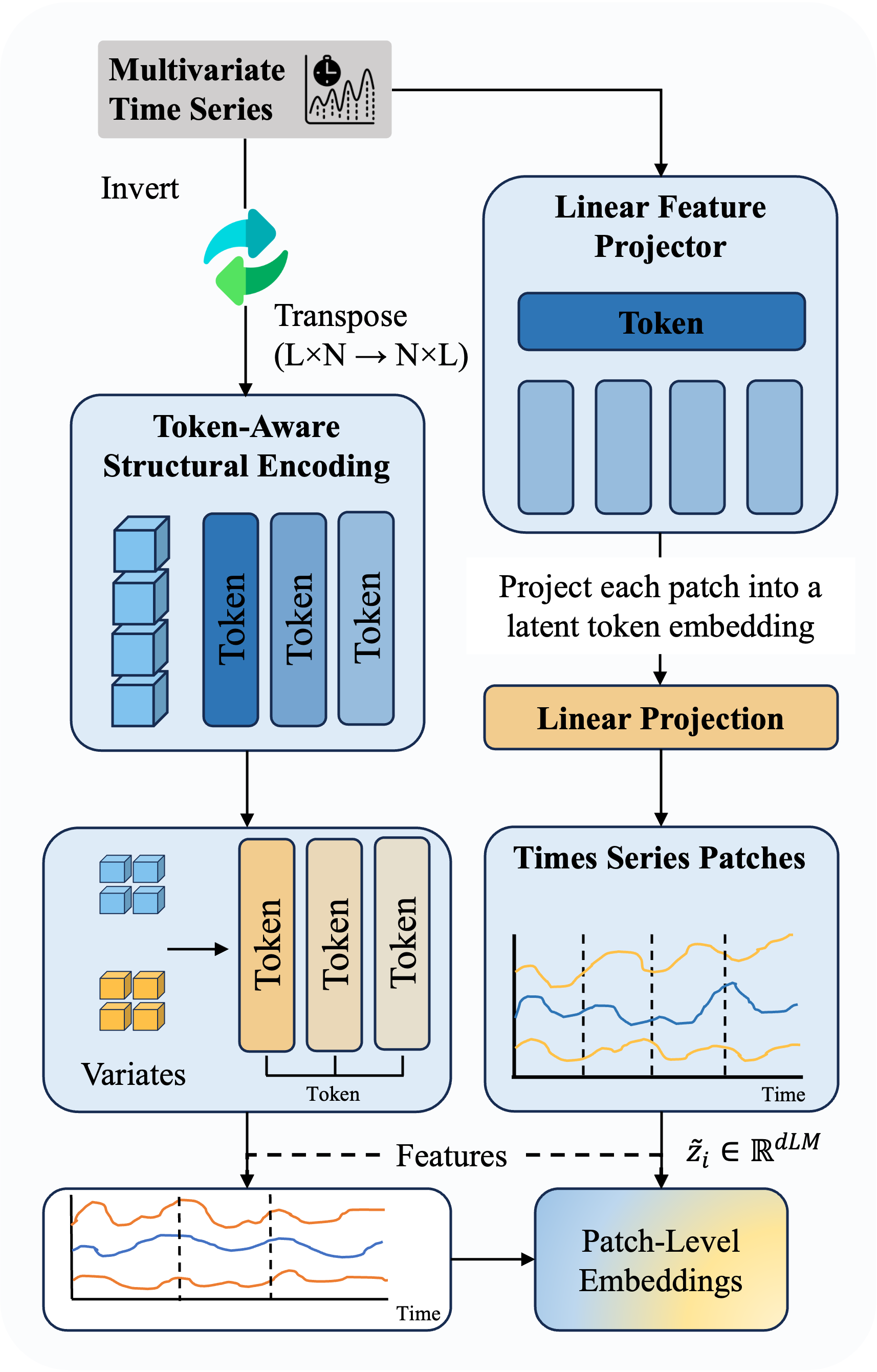}}
\caption{Token-aware structural encoding and patch projection process. The multivariate time series input is transposed to form variable-wise tokens, which are processed by a token-aware structural encoder to model inter-variable dependencies. The encoded representations are then segmented into fixed-length patches and projected into the embedding space expected by pretrained language models, forming aligned patch-level embeddings for semantic reprogramming.}
\label{fig02}
\end{figure}

\subsection{Token-Aware Structural Encoder}\label{ss1}
In conventional transformer-based time series models, temporal tokens are extracted along the time axis, where each token corresponds to a timestamp. However, this approach often entangles temporal dynamics with structural dependencies across variables, limiting the model's ability to generalize across variable-wise patterns. To address this, SEED adopts a token-aware structural encoder inspired by the inverted attention mechanism of iTransformer~\cite{liu2023itransformer}, which redefines each variable as a distinct token and performs attention along the variable dimension.

Formally, given a multivariate time series input $X \in \mathbb{R}^{L \times N}$, where $L$ denotes the sequence length and $N$ the number of variables, we first reshape the input to represent each variable as a token. This is achieved by transposing the input to obtain $X' \in \mathbb{R}^{N \times L}$, where each $x_i \in \mathbb{R}^{L}$ corresponds to the temporal evolution of the $i$-th variable. This allows the model to focus on learning structural dependencies among variables, rather than only temporal correlations.

The token-aware encoder comprises a stack of $M$ self-attention layers applied over the variable dimension. At each layer, the multi-head attention is computed as:
\begin{equation}
\mathrm{Attention}(Q, K, V) = \mathrm{softmax}\left(\frac{QK^\top}{\sqrt{d}}\right)V
\end{equation}
where $Q, K, V \in \mathbb{R}^{N \times d}$ are the query, key, and value matrices derived from linear projections of the variable-wise token embeddings, and $d$ is the embedding dimension.

To retain temporal information while focusing attention over variables, the input $X'$ is first processed by a temporal projection function $\phi: \mathbb{R}^{L} \to \mathbb{R}^{d}$ that maps each temporal trajectory into a latent space. This projection is implemented as a linear layer or 1D convolution:
\[x_i' = \phi(x_i), \quad \text{for } i = 1, \dots, N
\]
yielding a token matrix $T \in \mathbb{R}^{N \times d}$ that serves as the input to the variable-attention layers.

Residual connections and layer normalization are applied to ensure stability during training. Let $M$ denote the number of stacked attention layers. At each layer $l = 1, \dots, M$, the variable-token embeddings are updated via residual attention and feedforward updates:
\[
T^{(l+1)} = \mathrm{LayerNorm}\left(T^{(l)} + \mathrm{Attention}^{(l)}(T^{(l)})\right)
\]
\[
T^{(l+1)} = \mathrm{LayerNorm}\left(T^{(l+1)} + \mathrm{FFN}^{(l)}(T^{(l+1)})\right)
\]
where $\mathrm{FFN}^{(l)}(\cdot)$ denotes a two-layer feedforward network with ReLU activation.

This variable-wise attention formulation enables the encoder to learn high-order structural correlations across variables, making it well-suited for multivariate tasks. The resulting token representations serve as structure-aware features for subsequent patch projection and semantic alignment stages in SEED. An overview of this mechanism is also reflected in Fig.~\ref{fig02}.

\subsection{Patch Projection and Alignment}\label{ss2}
While the token-aware encoder in Section~\ref{ss1} learns variable-wise structural representations, these representations must be transformed into a token format compatible with large language models (LLMs) to enable semantic reasoning. This section introduces a patch-based projection module that bridges the structure-semantic gap by reshaping numerical representations into language-model-friendly embeddings.

Let the output of the encoder be denoted as $\mathbf{T} \in \mathbb{R}^{N \times D \times L}$, where $N$ is the number of variables (tokens), $D$ is the hidden dimension per variable, and $L$ is the temporal length. We first transpose and reshape $\mathbf{T}$ into a unified temporal layout:
\begin{equation}
\mathbf{E} = \text{Reshape}(\mathbf{T}) \in \mathbb{R}^{L \times (N \cdot D)},
\end{equation}
where each row $\mathbf{E}_t \in \mathbb{R}^{N \cdot D}$ represents the structural embedding of all variables at time step $t$.

To encode local temporal context, we divide $\mathbf{E}$ into non-overlapping patches of length $P$. Each patch is represented as:
\[
\mathbf{p}_i = \{\mathbf{E}_{(i-1)P + 1}, \ldots, \mathbf{E}_{iP}\} \in \mathbb{R}^{P \times (N \cdot D)}, \quad i = 1, \ldots, \left\lfloor \frac{L}{P} \right\rfloor,
\]
where $P$ is the patch size, and $i$ indexes the patch along the temporal axis. Each patch $\mathbf{p}_i$ captures a short-range temporal evolution across all variables.

To facilitate processing by pretrained LLMs, we project each patch into a latent token embedding $\mathbf{z}_i \in \mathbb{R}^{d_{\text{LM}}}$, where $d_{\text{LM}}$ is the target embedding dimension expected by the LLM:
\begin{equation}
\mathbf{z}_i = \mathbf{W}_p \cdot \text{Flatten}(\mathbf{p}_i) + \mathbf{b}_p,
\end{equation}
where $\mathbf{W}_p \in \mathbb{R}^{d_{\text{LM}} \times (P \cdot N \cdot D)}$ is a learnable weight matrix, and $\mathbf{b}_p \in \mathbb{R}^{d_{\text{LM}}}$ is the bias vector. The flatten operation converts each patch from a matrix to a vector in $\mathbb{R}^{P \cdot N \cdot D}$.

To encode relative position within the sequence, we introduce a learnable temporal positional embedding $\mathbf{e}_i^{\text{pos}} \in \mathbb{R}^{d_{\text{LM}}}$ for each patch:
\begin{equation}
\tilde{\mathbf{z}}_i = \mathbf{z}_i + \mathbf{e}_i^{\text{pos}}.
\end{equation}

The output of this stage is a sequence of projected tokens:
\[
\mathbf{Z} = \{\tilde{\mathbf{z}}_1, \tilde{\mathbf{z}}_2, \ldots, \tilde{\mathbf{z}}_M\} \in \mathbb{R}^{M \times d_{\text{LM}}},
\]
where $M = \left\lfloor \frac{L}{P} \right\rfloor$ is the number of patches. This sequence $\mathbf{Z}$ serves as the input to the semantic reprogramming module described in Section~\ref{ss3}.

Unlike traditional time series models that directly feed temporal vectors into decoders, our patch projection module enables a twofold innovation: (1) it preserves structural correlations by aggregating patch-wise multivariable embeddings from the encoder; (2) it aligns token embeddings with pretrained LLM input formats without retraining, enabling a seamless interface between structural encoding and semantic inference. This patch embedding interface forms the structural-semantic bridge with LLMs, as visualized in the right half of Fig.~\ref{fig02}.
\begin{figure}[b]
\centerline{\includegraphics[width=0.9\columnwidth]{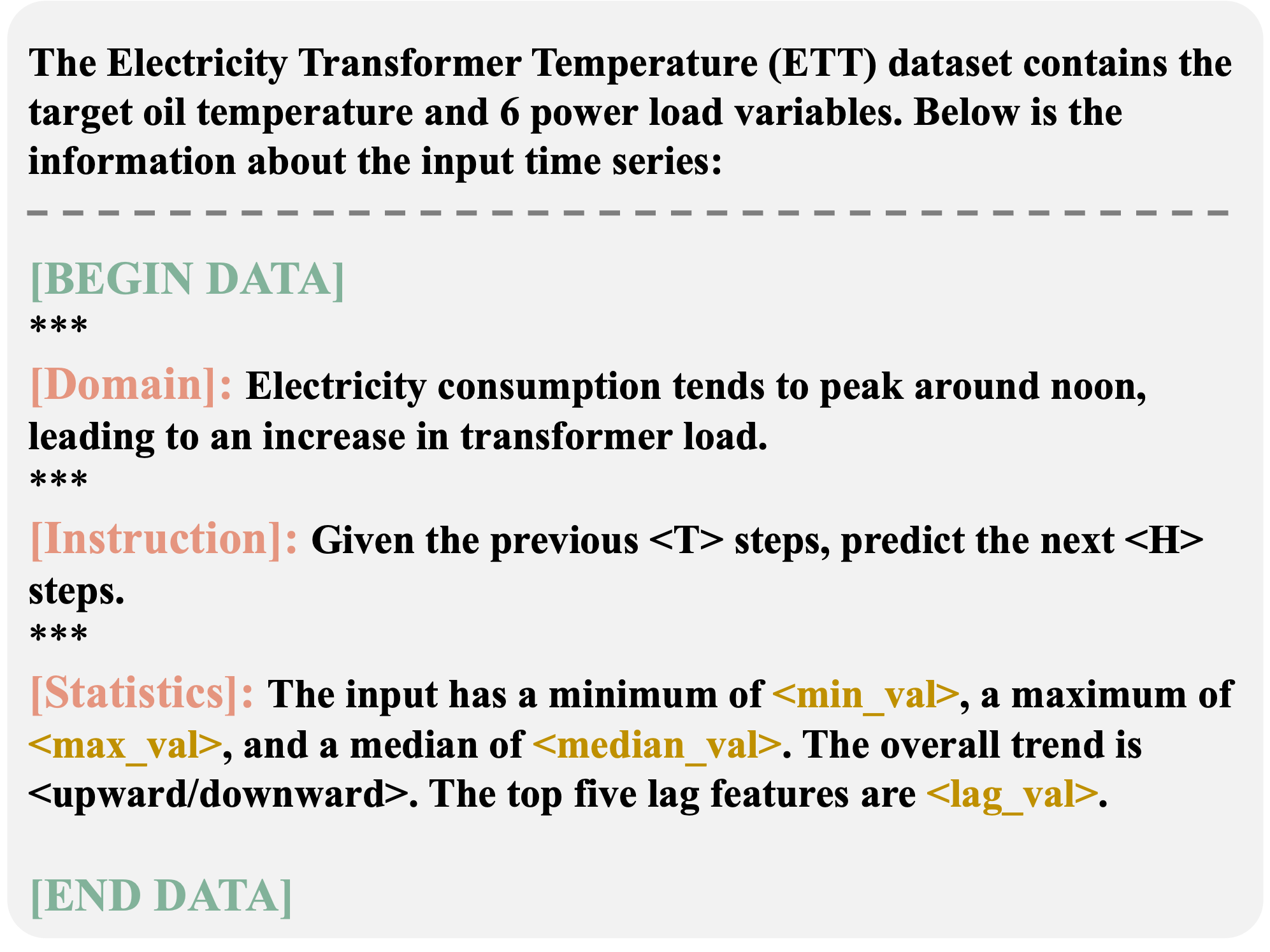}}
\caption{Prompt construction example. The structured prompt includes domain knowledge, task-specific instructions, and statistical features dynamically derived from the input series, which guides the frozen LLMs during inference.}
\label{fig:prompt}
\end{figure}

\subsection{Semantic Reprogramming via Prototypes and Prompts}\label{ss3}
While the projected patch tokens $\mathbf{Z} = \{\mathbf{z}_1, \mathbf{z}_2, \dots, \mathbf{z}_N\}$ from Section~\ref{ss2} are aligned in dimensionality with the target language model, they remain fundamentally numerical and lack semantic alignment with natural language representations. To bridge this gap, SEED introduces a semantic reprogramming mechanism inspired by prompt-tuning strategies in Time-LLM~\cite{jin2023time}, which maps numerical embeddings into interpretable token sequences through the use of learnable prototypes and task-aware prompts.

Specifically, let $\mathcal{P} = \{\mathbf{p}_1^t, \ldots, \mathbf{p}_K^t\} \subset \mathbb{R}^{d_{\text{LM}}}$ denote a set of $K$ learnable \emph{textual prototypes}, where each $\mathbf{p}_k^t$ corresponds to a semantic anchor associated with common temporal patterns or domain-relevant states. Given a projected patch token $\mathbf{z}_i \in \mathbb{R}^{d_{\text{LM}}}$, we compute its similarity to each prototype via scaled dot-product attention:
\begin{equation}
\alpha_{i,k} = \frac{\exp\left( \mathbf{z}_i^\top \mathbf{p}_k^t / \sqrt{d_{\text{LM}}} \right)}{\sum_{j=1}^K \exp\left( \mathbf{z}_i^\top \mathbf{p}_j^t / \sqrt{d_{\text{LM}}} \right)},
\end{equation}
where $\alpha_{i,k}$ denotes the normalized attention score between $\mathbf{z}_i$ and prototype $\mathbf{p}_k^t$.

We then synthesize a semantically grounded token $\tilde{\mathbf{z}}_i$ via a convex combination of prototypes:
\begin{equation}
\tilde{\mathbf{z}}_i = \sum_{k=1}^K \alpha_{i,k} \cdot \mathbf{p}_k^t,
\end{equation}
where $\tilde{\mathbf{z}}_i$ preserves the original temporal pattern of $\mathbf{z}_i$ while acquiring linguistic semantics from $\mathcal{P}$. This attention-based transformation of patch tokens into semantically grounded representations is visualized in Fig.~\ref{fig03}. Further, an example of the prompt format used in this stage is shown in Fig.~\ref{fig:prompt}, where domain context, instructions, and input statistics are assembled to construct task-aware guidance.

To further guide the LLM decoder toward task-specific behavior, we prepend a task prompt sequence $\mathcal{T} = \{\mathbf{t}_1, \ldots, \mathbf{t}_L\} \subset \mathbb{R}^{d_{\text{LM}}}$, where each $\mathbf{t}_\ell$ is a learned embedding associated with a particular downstream task (e.g., forecasting, imputation, anomaly detection). The final reprogrammed input to the language model becomes:
\begin{equation}
\mathbf{X}^{\text{LM}} = [\mathbf{t}_1, \ldots, \mathbf{t}_L, \tilde{\mathbf{z}}_1, \ldots, \tilde{\mathbf{z}}_N] \in \mathbb{R}^{(L + N) \times d_{\text{LM}}},
\end{equation}
where $[\cdot]$ denotes sequence concatenation along the temporal axis.

Through this mechanism, SEED transforms numerical patches into semantic tokens aligned with the language model’s input space. By grounding variable dynamics with learnable prototypes and steering predictions via task prompts, SEED forms a lightweight semantic interface that enables the frozen LLM to perform inference without fine-tuning.

\subsection{Decoding via Frozen LLMs}\label{ss4}
In the final stage of SEED, the semantically reprogrammed token sequence is passed to a frozen LLM for time series prediction. Unlike conventional approaches that fine-tune LLMs for downstream tasks, our framework treats the LLM as a zero-shot reasoning engine, using only prompt-based interaction to guide the prediction behavior.

In practice, the frozen LLM can be instantiated using any autoregressive decoder-only model such as GPT-2~\cite{radford2019language} or LLaMA-2~\cite{touvron2023llama}, depending on deployment requirements. Since SEED decouples structural encoding from semantic inference, the framework remains agnostic to the specific LLM used, enabling flexible integration without altering its architecture.

\begin{figure}[t]
\centerline{\includegraphics[width=1\columnwidth]{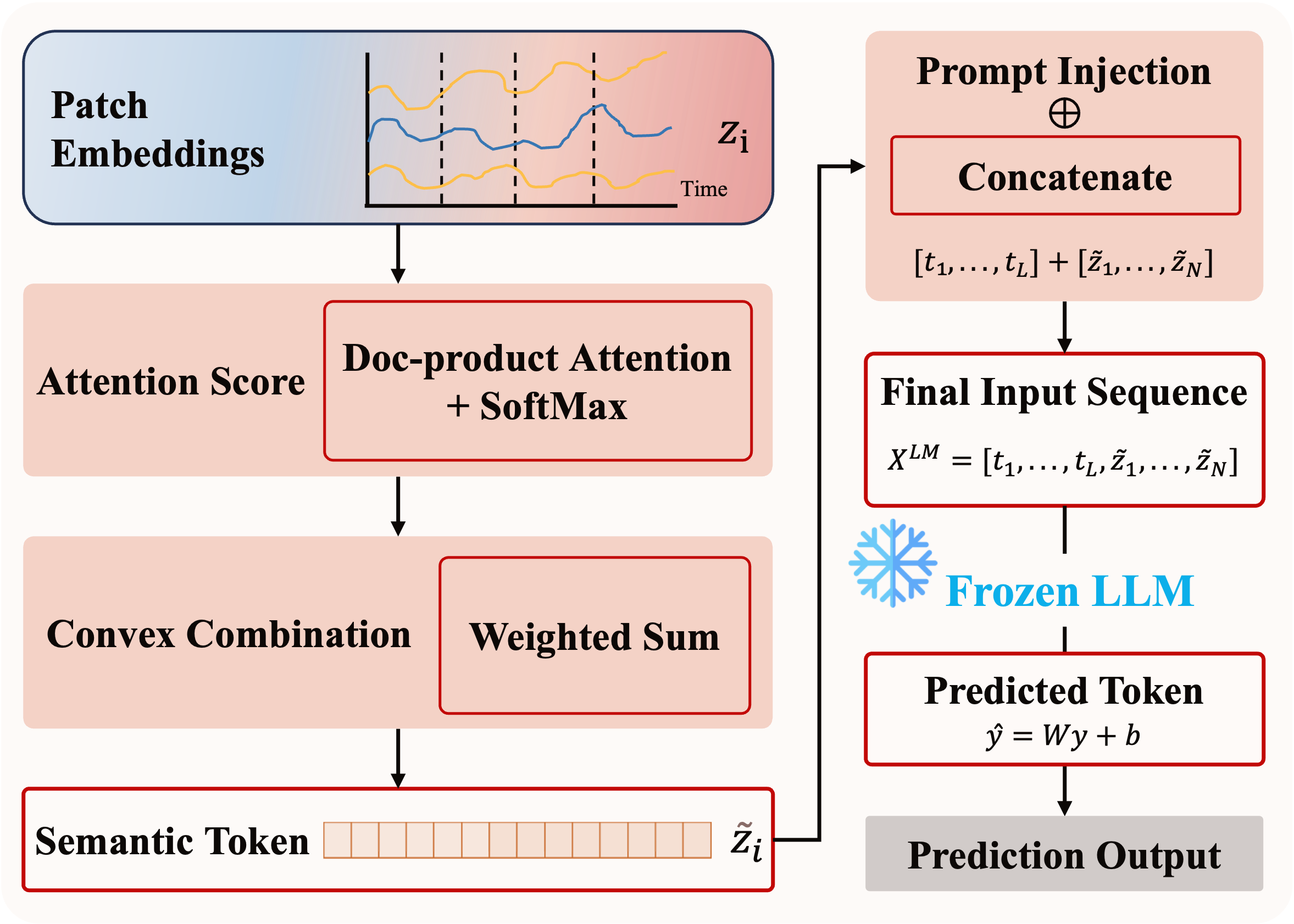}}
\caption{Semantic reprogramming and decoding with a frozen language model. Projected patch embeddings $\mathbf{z}_i$ are semantically grounded via scaled dot-product attention with learnable textual prototypes. The resulting semantic tokens $\tilde{\mathbf{z}}_i$ are concatenated with task prompts to form the final input sequence $X^{\text{LM}}$, which is then processed by a frozen autoregressive decoder. The predicted token is mapped back to the original value space via a linear head to generate the predicted value.}
\label{fig03}
\end{figure}

Formally, let $\mathbf{Z} = \{\mathbf{z}_1, \mathbf{z}_2, \dots, \mathbf{z}_N\} \in \mathbb{R}^{N \times d_{\text{LM}}}$ denote the sequence of semantic embeddings obtained from the prototype-based reprogramming module, where $N$ is the number of patch-level tokens and $d_{\text{LM}}$ is the embedding dimension of the LLM. This sequence is prepended with a task-specific prompt $\mathbf{p}_{\text{task}} \in \mathbb{R}^{1 \times d_{\text{LM}}}$ that instructs the LLM on the type of temporal reasoning to perform. The complete decoding process from semantic tokens to predicted values is shown in Fig.~\ref{fig03}; and the final input sequence becomes:
\[
\mathbf{Z}' = \left[\mathbf{p}_{\text{task}}; \mathbf{z}_1; \mathbf{z}_2; \ldots; \mathbf{z}_N \right] \in \mathbb{R}^{(N+1) \times d_{\text{LM}}}
\]

The frozen LLM processes $\mathbf{Z}'$ and auto-regressively generates a predicted token $\hat{\mathbf{y}} \in \mathbb{R}^{d_{\text{LM}}}$, which is then mapped back to the original value space via a learned linear head:
\[
\hat{y} = \mathbf{W}_{\text{out}} \cdot \hat{\mathbf{y}} + \mathbf{b}_{\text{out}}, \quad \mathbf{W}_{\text{out}} \in \mathbb{R}^{1 \times d_{\text{LM}}},\ \mathbf{b}_{\text{out}} \in \mathbb{R}
\]

This setup enables SEED to preserve the pretrained capabilities of the LLM while adapting it to time series domains through modular embedding alignment. Crucially, the language model remains frozen, and all task adaptation is achieved through the upstream encoder, patch projection, and semantic reprogramming components. This design not only reduces training overhead but also improves generalization across heterogeneous forecasting tasks.

In summary, SEED leverages the structural awareness of token-wise attention, the alignment power of patch embedding, and the reasoning ability of pretrained LLMs to perform robust and extensible time series prediction. This lightweight integration enables effective deployment across forecasting scenarios without task-specific retraining or fine-tuning of the LLM.

\begin{table*}[t]
\renewcommand{\arraystretch}{1.6}  
\centering
\caption{Multivariate forecasting comparison on benchmark datasets.}
\label{tab:comparison}
\begin{tabular}{@{}l|cc|cc|cc|cc|cc|cc@{}}
\toprule
\multirow{2}{*}{\textbf{Dataset}} & \multicolumn{2}{c|}{\textbf{SEED (Ours)}} 
& \multicolumn{2}{c|}{\textbf{Time-LLM~\cite{jin2023time}}} 
& \multicolumn{2}{c|}{\textbf{iTransformer~\cite{liu2023itransformer}}} 
& \multicolumn{2}{c|}{\textbf{Autoformer~\cite{wu2021autoformer}}} 
& \multicolumn{2}{c|}{\textbf{Informer~\cite{zhou2021informer}}} 
& \multicolumn{2}{c}{\textbf{Reformer~\cite{kitaev2020reformer}}} \\
\cmidrule(lr){2-3} \cmidrule(lr){4-5} \cmidrule(lr){6-7} \cmidrule(lr){8-9} \cmidrule(lr){10-11} \cmidrule(lr){12-13}
& MSE ↓ & MAE ↓ & MSE ↓ & MAE ↓ & MSE ↓ & MAE ↓ & MSE ↓ & MAE ↓ & MSE ↓ & MAE ↓ & MSE ↓ & MAE ↓ \\
\midrule
\textbf{ETTh1} & 0.413 & 0.430 & \textbf{0.408} & \textbf{0.423} & 0.496 & 0.487 & 0.722 & 0.598 & 1.225 & 0.817 & 1.241 & 0.835 \\
\textbf{ETTh2} & \textbf{0.330} & \textbf{0.379} & 0.334 & 0.383 & 0.450 & 0.459 & 0.441 & 0.457 & 3.922 & 1.653 & 3.527 & 1.472 \\
\textbf{ETTm1} & \textbf{0.329} & \textbf{0.372} & 0.388 & 0.403 & 0.588 & 0.517 & 0.620 & 0.517 & 1.163 & 0.791 & 1.264 & 0.826 \\
\textbf{ETTm2} & 0.285 & \textbf{0.333} & \textbf{0.284} & 0.339 & 0.327 & 0.388 & 0.433 & 0.371 & 3.658 & 1.489 & 3.581 & 1.487 \\
\textbf{Weather} & \textbf{0.225} & \textbf{0.257} & 0.237 & 0.270 & 0.338 & 0.310 & 0.353 & 0.382 & 0.584 & 0.527 & 0.447 & 0.453 \\
\textbf{ECL} & \textbf{0.158} & \textbf{0.252} & 0.167 & 0.263 & 0.178 & 0.346 & 0.404 & 0.338 & 1.281 & 0.929 & 1.289 & 0.904 \\
\textbf{Traffic} & 0.388 & 0.264 & \textbf{0.390} & \textbf{0.263} & 0.428 & 0.282 & 0.833 & 0.502 & 1.591 & 0.832 & 1.618 & 0.851 \\
\textbf{Solar-Energy} & 0.243 & 0.263 & 0.270 & 0.307 & \textbf{0.233} & \textbf{0.262} & 0.866 & 0.491 & 1.137 & 0.744 & 1.724 & 0.945 \\
\bottomrule
\end{tabular}
\end{table*}

\section{Experiments and Results}
To evaluate the effectiveness of SEED in multivariate time series prediction, all experiments are conducted on a dedicated computing server running Ubuntu 20.04 with an NVIDIA RTX 4090 GPU (24 GB memory), CUDA 11.8, and Python 3.8 (Miniconda). The models are implemented using PyTorch and trained with the Adam optimizer. Unless otherwise specified, all encoders are initialized with the iTransformer backbone, and the decoder adopts a frozen LLaMA-2 or GPT-2 model via Hugging Face Transformers. Each model is trained with early stopping based on validation loss, using a fixed random seed to ensure reproducibility. The default patch length $P$ is set to 16, and the prototype set contains $K=8$ learnable vectors unless stated otherwise. The batch size and learning rate are tuned separately for each dataset, with all models trained under identical conditions for fair comparison.

We benchmark SEED on eight widely used multivariate time series datasets: ETTh1, ETTh2, ETTm1, ETTm2, Weather, Electricity (ECL), Traffic, and Solar Energy. These datasets represent diverse temporal domains, including meteorological records, electrical consumption, road occupancy, and solar production, with varying sequence lengths and sampling frequencies. ETTh and ETTm consist of hourly and minute-level readings of power-related variables; Weather and ECL contain regional climate and electricity usage data respectively; Traffic captures hourly occupancy rates of road sensors; and Solar records photovoltaic power generation over time. For each dataset, we follow standard train-validation-test splits and apply z-score normalization over the training set. The forecasting task aims to predict multivariate future trajectories given a fixed-length historical window, evaluating the framework’s generalization ability across heterogeneous temporal patterns.

Table~\ref{tab:comparison} reports the multivariate forecasting performance of our proposed SEED framework in comparison with five representative baselines: Time-LLM~\cite{jin2023time}, iTransformer~\cite{liu2023itransformer}, Autoformer~\cite{wu2021autoformer}, Informer~\cite{zhou2021informer}, and Reformer~\cite{kitaev2020reformer}, across eight publicly available datasets. SEED achieves the best overall results on five datasets with respect to both MSE and MAE, including ETTh2, ETTm1, Weather, ECL, and Solar-Energy.

On ETTh2, SEED achieves an MSE of 0.330 and MAE of 0.379, outperforming all baselines, including Time-LLM (MSE: 0.334, MAE: 0.383) and iTransformer (MSE: 0.450, MAE: 0.459). On Weather and ECL, SEED obtains the lowest MSE scores of 0.225 and 0.158 respectively, indicating a strong capability to capture temporal and inter-variable dependencies in high-dimensional climate and energy usage patterns. On ETTm1, SEED outperforms all competing methods with the lowest MAE of 0.372 and ties for the best MSE, confirming its stability in minute-level high-frequency settings. On Solar-Energy, SEED performs comparably with iTransformer in terms of MSE, both achieving approximately 0.24, while maintaining a lower MAE. This validates the effectiveness of the structural-semantic decoupling of SEED.

Compared with Informer and Reformer, SEED consistently yields substantial improvements across all datasets. For instance, on ETTh2, SEED reduces MSE from Informer's 3.922 and Reformer's 3.527 down to 0.330, achieving over a 10-fold performance gain. These results highlight the advantage of SEED modular design, which enables generalization across heterogeneous time series domains by aligning structural encodings with language model semantics.

On Solar-Energy, SEED matches the best MSE score of 0.233 and retains the lowest MAE among all models, indicating strong alignment between structural representations and LLM-based semantic reasoning. Furthermore, SEED significantly outperforms Informer and Reformer on all datasets, with particularly large improvements observed on ETTh2 and ETTm2, where SEED reduces MSE by over 4.0 times compared to Informer, highlighting its superior robustness in long-horizon forecasting scenarios.

In contrast, SEED performs slightly behind Time-LLM on Traffic and ETTh1 in terms of MSE. This performance gap can be attributed to the presence of extreme outliers and long-tail error distributions in traffic flow and industrial load datasets. Due to the use of a frozen LLM decoder without task-specific fine-tuning, SEED may be less effective in capturing subtle signal distortions or sparse events. Nevertheless, SEED still achieves the lowest MAE on Traffic, suggesting more reliable point-wise predictions even under volatile input conditions.

These results demonstrate the versatility and effectiveness of SEED in multivariate time series forecasting. By separating structural representation learning from downstream inference through semantic reprogramming, SEED establishes a unified and extensible architecture capable of adapting to a wide range of temporal tasks while leveraging the generalization strength of pretrained LLMs.

\section{Conclusion}
This paper presented SEED, a structural encoder for embedding-driven decoding, designed to bridge the gap between structural representation learning and semantic inference in multivariate time series forecasting. By integrating token-aware encoding, patch projection, semantic reprogramming, and frozen LLM decoding, SEED decouples numerical pattern modeling from task-specific reasoning. Experimental results on diverse temporal datasets demonstrated that SEED achieves competitive or superior performance compared to both traditional transformer variants and recent LLM-based frameworks. In particular, the combination of variable-wise structural attention and prototype-guided prompt alignment enables SEED to generalize across heterogeneous domains while maintaining lightweight inference. These findings support the potential of modular structural-semantic integration for advancing unified, transferable time series prediction systems.

\section*{Acknowledgment}
This research is supported by the Natural Science Foundation of China (62472361), the Suzhou Science and Technology Project-Key Industrial Technology Innovation (SYG202122), 2024 Suzhou Innovation Consortium Construction Project (LHT202406 ), Suzhou Municipal Key Laboratory for Intelligent Virtual Engineering (SZS2022004), the XJTLU Postgraduate Research Scholarship (Grand No. PGRS1906004), the  XJTLU AI University Research Centre, Zooming New Energy-XJTLU Smart Energy Joint Laboratory, Jiangsu Province Engineering Research Centre of Data Science and Cognitive Computation, Suzhou Data Innovation Application Laboratory and SIP High level innovation platform.

\bibliographystyle{IEEEtran}
\bibliography{ref}

\begin{thebibliography}{10}
\providecommand{\url}[1]{#1}
\csname url@samestyle\endcsname
\providecommand{\newblock}{\relax}
\providecommand{\bibinfo}[2]{#2}
\providecommand{\BIBentrySTDinterwordspacing}{\spaceskip=0pt\relax}
\providecommand{\BIBentryALTinterwordstretchfactor}{4}
\providecommand{\BIBentryALTinterwordspacing}{\spaceskip=\fontdimen2\font plus
\BIBentryALTinterwordstretchfactor\fontdimen3\font minus \fontdimen4\font\relax}
\providecommand{\BIBforeignlanguage}[2]{{%
\expandafter\ifx\csname l@#1\endcsname\relax
\typeout{** WARNING: IEEEtran.bst: No hyphenation pattern has been}%
\typeout{** loaded for the language `#1'. Using the pattern for}%
\typeout{** the default language instead.}%
\else
\language=\csname l@#1\endcsname
\fi
#2}}
\providecommand{\BIBdecl}{\relax}
\BIBdecl

\bibitem{mendis2024multivariate}
K.~Mendis, M.~Wickramasinghe, and P.~Marasinghe, ``Multivariate time series forecasting: A review,'' in \emph{Proceedings of the 2024 2nd Asia Conference on Computer Vision, Image Processing and Pattern Recognition}, 2024, pp. 1--9.

\bibitem{cai2024msgnet}
W.~Cai, Y.~Liang, X.~Liu, J.~Feng, and Y.~Wu, ``Msgnet: Learning multi-scale inter-series correlations for multivariate time series forecasting,'' in \emph{Proceedings of the AAAI Conference on Artificial Intelligence}, vol.~38, no.~10, 2024, pp. 11\,141--11\,149.

\bibitem{shao2024exploring}
Z.~Shao, F.~Wang, Y.~Xu, W.~Wei, C.~Yu, Z.~Zhang, D.~Yao, T.~Sun, G.~Jin, X.~Cao \emph{et~al.}, ``Exploring progress in multivariate time series forecasting: Comprehensive benchmarking and heterogeneity analysis,'' \emph{IEEE Transactions on Knowledge and Data Engineering}, 2024.

\bibitem{ray2023arima}
S.~Ray, A.~Lama, P.~Mishra, T.~Biswas, S.~S. Das, and B.~Gurung, ``An arima-lstm model for predicting volatile agricultural price series with random forest technique,'' \emph{Applied Soft Computing}, vol. 149, p. 110939, 2023.

\bibitem{chia2022long}
M.~Y. Chia, Y.~F. Huang, C.~H. Koo, J.~L. Ng, A.~N. Ahmed, and A.~El-Shafie, ``Long-term forecasting of monthly mean reference evapotranspiration using deep neural network: A comparison of training strategies and approaches,'' \emph{Applied Soft Computing}, vol. 126, p. 109221, 2022.

\bibitem{kumar2022multi}
R.~Kumar, P.~Kumar, and Y.~Kumar, ``Multi-step time series analysis and forecasting strategy using arima and evolutionary algorithms,'' \emph{International Journal of Information Technology}, vol.~14, no.~1, pp. 359--373, 2022.

\bibitem{lstm}
S.~Hochreiter and J.~Schmidhuber, ``Long short-term memory,'' \emph{Neural Comput.}, vol.~9, no.~8, p. 1735–1780, Nov. 1997.

\bibitem{vaswani2017attention}
A.~Vaswani, N.~Shazeer, N.~Parmar, J.~Uszkoreit, L.~Jones, A.~N. Gomez, {\L}.~Kaiser, and I.~Polosukhin, ``Attention is all you need,'' \emph{Advances in neural information processing systems}, vol.~30, 2017.

\bibitem{kitaev2020reformer}
N.~Kitaev, {\L}.~Kaiser, and A.~Levskaya, ``Reformer: The efficient transformer,'' in \emph{International Conference on Learning Representations (ICLR)}, 2020.

\bibitem{feng2024latent}
S.~Feng, C.~Miao, Z.~Zhang, and P.~Zhao, ``Latent diffusion transformer for probabilistic time series forecasting,'' in \emph{Proceedings of the AAAI Conference on Artificial Intelligence}, vol.~38, no.~11, 2024, pp. 11\,979--11\,987.

\bibitem{zhang2024multi}
Y.~Zhang, L.~Ma, S.~Pal, Y.~Zhang, and M.~Coates, ``Multi-resolution time-series transformer for long-term forecasting,'' in \emph{International conference on artificial intelligence and statistics}.\hskip 1em plus 0.5em minus 0.4em\relax PMLR, 2024, pp. 4222--4230.

\bibitem{kang2024transformer}
H.~Kang and P.~Kang, ``Transformer-based multivariate time series anomaly detection using inter-variable attention mechanism,'' \emph{Knowledge-Based Systems}, vol. 290, p. 111507, 2024.

\bibitem{zhou2021informer}
H.~Zhou, S.~Zhang, J.~Peng, S.~Zhang, J.~Li, H.~Xiong, and W.~Zhang, ``Informer: Beyond efficient transformer for long sequence time-series forecasting,'' in \emph{Proceedings of the AAAI conference on artificial intelligence}, vol.~35, no.~12, 2021, pp. 11\,106--11\,115.

\bibitem{wu2021autoformer}
H.~Wu, J.~Xu, J.~Wang, and M.~Long, ``Autoformer: Decomposition transformers with auto-correlation for long-term series forecasting,'' \emph{Advances in neural information processing systems}, vol.~34, pp. 22\,419--22\,430, 2021.

\bibitem{Yuqietal-2023-PatchTST}
Y.~Nie, N.~H.~Nguyen, P.~Sinthong, and J.~Kalagnanam, ``A time series is worth 64 words: Long-term forecasting with transformers,'' in \emph{International Conference on Learning Representations (ICLR)}, 2023.

\bibitem{liu2023itransformer}
Y.~Liu, T.~Hu, H.~Zhang, H.~Wu, S.~Wang, L.~Ma, and M.~Long, ``itransformer: Inverted transformers are effective for time series forecasting,'' in \emph{International Conference on Learning Representations (ICLR)}, 2024.

\bibitem{xue2023promptcast}
H.~Xue and F.~D. Salim, ``Promptcast: A new prompt-based learning paradigm for time series forecasting,'' \emph{IEEE Transactions on Knowledge and Data Engineering}, vol.~36, no.~11, pp. 6851--6864, 2023.

\bibitem{jin2023time}
M.~Jin, S.~Wang, L.~Ma, Z.~Chu, J.~Y. Zhang, X.~Shi, P.-Y. Chen, Y.~Liang, Y.-F. Li, S.~Pan, and Q.~Wen, ``{Time-LLM}: Time series forecasting by reprogramming large language models,'' in \emph{International Conference on Learning Representations (ICLR)}, 2024.

\bibitem{garza2023timegpt}
A.~Garza, C.~Challu, and M.~Mergenthaler-Canseco, ``Timegpt-1,'' \emph{arXiv preprint arXiv:2310.03589}, 2023.

\bibitem{radford2019language}
A.~Radford, J.~Wu, R.~Child, D.~Luan, D.~Amodei, I.~Sutskever \emph{et~al.}, ``Language models are unsupervised multitask learners,'' \emph{OpenAI blog}, vol.~1, no.~8, p.~9, 2019.

\bibitem{touvron2023llama}
H.~Touvron, L.~Martin, K.~Stone, P.~Albert, A.~Almahairi, Y.~Babaei, N.~Bashlykov, S.~Batra, P.~Bhargava, S.~Bhosale \emph{et~al.}, ``Llama 2: Open foundation and fine-tuned chat models,'' \emph{arXiv preprint arXiv:2307.09288}, 2023.

\end{thebibliography}




\end{document}